\documentclass[sigconf]{acmart}

\usepackage{booktabs}
\usepackage[ruled,vlined]{algorithm2e}
\usepackage[table]{xcolor}
\usepackage{comment}
\usepackage{xcolor}
\usepackage{colortbl}
\SetKwComment{tcp}{$\blacktriangleright$}{}
\usepackage{arydshln}   
\usepackage{graphicx}
\usepackage{multirow}
\usepackage{subcaption}
\usepackage{adjustbox}
\usepackage{changepage} 
\usepackage{setspace}   

\AtBeginDocument{%
  }

\acmConference[SIGIR '26]
{July 2026}
{Melbourne, Australia}

\acmYear{2026}
\copyrightyear{2026}

\begin{document}

\title{Learning to Reason for Multi-Step Retrieval of Personal Context in Personalized Question Answering}

\author{Maryam Amirizaniani}
\authornote{Work done while visiting CIIR at UMass Amherst.}
\email{amaryam@uw.edu}
\affiliation{
  \institution{University of Washington}
  \city{Seattle}
  \state{WA}
  \country{USA}
}

\author{Alireza Salemi}
\email{asalemi@cs.umass.edu}
\affiliation{
  \institution{University of Massachusetts Amherst}
  \city{Amherst}
  \state{MA}
  \country{USA}
}

\author{Hamed Zamani}
\email{zamani@cs.umass.edu}
\affiliation{
  \institution{University of Massachusetts Amherst}
  \city{Amherst}
  \state{MA}
  \country{USA}
}

\begin{abstract}
Personalization in Question Answering (QA) requires answers that are both accurate and aligned with users' background, preferences, and historical context. Existing state-of-the-art methods primarily rely on retrieval-augmented generation (RAG) solutions that construct personal context by retrieving relevant items from the user's profile. Existing methods use the user's query directly to retrieve personal documents and such strategies often lead to surface-level personalization. 
We propose PR$^{2}$ (\textbf{P}ersonalized \textbf{R}etrieval-Augmented \textbf{R}easoning), a reinforcement learning framework that integrates reasoning and retrieval from personal context for personalization. PR$^{2}$ learns adaptive retrieval-reasoning policies, determining when to retrieve, what evidence to retrieve from user profiles, and how to incorporate it into intermediate reasoning steps. By optimizing multi-turn reasoning trajectories under a personalized reward function, the framework reinforces reasoning paths that better align with user-specific preferences and contextual signals reflected by the reward model. Extensive experiments on the LaMP-QA benchmark using three LLMs show that PR$^{2}$ consistently outperforms strong baselines, achieving an average relative improvement of 8.8\%-12\% in personalized QA. 
\end{abstract}
\maketitle


\section{Introduction}
Personalization in LLMs has become increasingly important, particularly in information-seeking applications such as recommendation, search, and Question Answering (QA)~\cite{wu2025rlpf, tsai-etal-2024-leveraging, bismay-etal-2025-reasoningrec, cho-etal-2025-modeling, sun-etal-2025-persona}. Among these, personalized text generation for question answering (QA) remains underexplored. In personalized QA, models must generate responses that are tailored to users' preferences and contextual information~\cite{salemi2025pathways, salemi-zamani-2025-lamp, kumar2024longlamp}. For example, answering \textit{``What's the best way to lose weight?''} requires accounting for personalized factors such as dietary restrictions, activity level, and medical history. The same question asked by different users may require different answers; identical responses suggest a failure to achieve personalization.

State-of-the-art personalized QA methods adopt a standard retrieval-augmented generation (RAG) paradigm, where personal context  is retrieved from the user profile using the user's question as the search query and appended to the input prompt to guide response generation~\cite{10.1145/3626772.3657783, 10.1145/3626772.3657957, 10.1145/3746252.3760851, islam-etal-2024-open}. \textit{We argue that effective personalization can be achieved by identifying potential personalized aspects that would impact the generated answer. Identifying these aspects require accurate reasoning and sometimes generation of multiple queries that explore different personalization aspects.} For instance, to answer the above mentioned question, it is important to identify the user's occupation to estimate the level of activity in their lifestyle and an effective model may generate the query ``occupation'' or later the query ``age'' to provide an accurate personalized answer. This demonstrates the importance of both reasoning and multi-step querying and retrieval from the user profile.

Inspired by recent work on retrieval-augmented reasoning (RAR) methods for question answering and deep research agents~\cite{jin2025searchr, tran-etal-2025-rare, soudani2025uncertainty, jin2025longcontext, li-etal-2025-search}, 
%
we propose PR$^{2}$ (\textbf{P}ersonalized \textbf{R}etrieval-Augmented \textbf{R}easoning), a reinforcement learning (RL)~\cite{sutton1998reinforcement} framework that embeds retrieval into the reasoning process to enable adaptive and user-specific personalization. To optimize this integrated retrieval–reasoning process, PR$^{2}$ is trained with Group Relative Policy Optimization (GRPO)~\cite{shao2024deepseekmath, guo2025deepseek}, which models retrieval as part of the environment and optimizes full generation trajectories under personalized reward signals. During training, PR$^{2}$ samples both personalized and non-personalized trajectories to stabilize RL optimization and favoring user-aligned reasoning. The model supports multi-turn retrieval and reasoning, dynamically deciding when to retrieve, what information to retrieve, when to stop for generating the response, and how to incorporate it by interleaving retrieval actions with reasoning steps. 
This unified design enables adaptive and iterative retrieval and consumption of personal data throughout the reasoning process. Experiments on LaMP-QA~\cite{salemi-zamani-2025-lamp}, a recent personalized long-form question answering benchmark consisting of three diverse datasets, show that PR$^{2}$ consistently outperforms strong baselines, achieving relative improvements of 8.8\%-12.0\% across different LLMs, highlighting its effectiveness for personalized QA. To improve reproducibility of this research, we open-source our implementation at \url{https://github.com/maryamamiri114/PR2}.


\section{Related Work}
\noindent {\textbf{Reasoning in LLMs.}}
Many studies investigate methods to improve LLM reasoning across a wide range of tasks ~\cite{aggarwal2025l, ye2025limo, chen-etal-2025-towards-medical, 10.1145/3746252.3761120} including QA~\cite{jin2025searchr, 10.1145/3626772.3657783}. A common paradigm is step-by-step reasoning, often formalized as chain-of-thought (CoT)~\cite{wei2022chain}, which decomposes complex problems into intermediate steps. However, many methods rely on carefully designed prompts, limiting generalization. Recent work instead explores fine-tuning and RL to learn more robust reasoning behaviors~\cite{zhang-etal-2025-legal, zhang-etal-2025-beyond-guilt, lin2025recr, xie2025logic, 10.1145/3731120.3744598}.

\medskip
{\noindent{\textbf{Personalized QA.}}}
Personalized QA has largely focused on conditioning model behavior using user history at inference time~\cite{10.1145/3626772.3657957, 10.1145/3627673.3679832, kim2025ipqa}. A common strategy is In-Context Learning, where past user query–response pairs are inserted into the prompt to induce personalized behavior~\cite{10.1145/3604915.3610646, 10.1145/3616855.3635845, zhao2025do, zhao2025do, wang2023zero, salemi-etal-2024-lamp}. Other approaches directly utilize complete user histories for prompting~\cite{10.1145/3640457.3688161, liu-etal-2025-llms, kim-yang-2025-shot}, enabling personalization through structured profile representations rather than raw interaction logs. RAG extends this paradigm by retrieving semantically similar examples from user histories~\cite{10.1145/3626772.3657957, 10.1145/3637528.3671470, zhuang2024hydra}. Subsequent work improves retrieval using reward signals from LLM outputs or dynamically selecting retrieval strategies based on input~\cite{10.1145/3626772.3657783, salemi2025pathways}. Personalized reasoning in QA lies at the intersection of LLM reasoning and user-aware personalization. Early work relies on personal embeddings such as PPlug~\cite{liu-etal-2025-llms} for reasoning templates, while recent approaches use RL with contrastive rewards~\cite{10.1145/3746252.3760851}. However, existing work primarily improves retrieval to enhance personalized answers. To address this gap, we propose PR$^{2}$, which integrates retrieved user-specific evidence directly into the intermediate reasoning process to generate personalized responses.

\begin{figure}[t]
    \centering
    \includegraphics[scale=0.6]{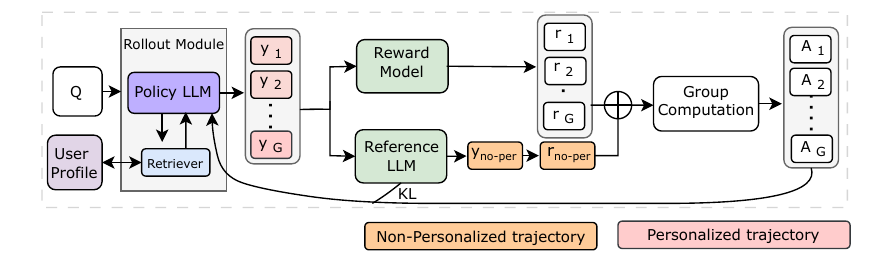}
    \vspace{-0.5em}
    \caption{An overview of PR$^{2}$ optimization based on GRPO.}
    \label{fig:framework}
\end{figure} 

\section{Methodology}
We present \textsc{PR$^{2}$}, a framework for training LLMs for personalized QA via retrieval-augmented reasoning with GRPO optimization, enabling the model to learn how to effectively retrieve personal information from user profiles. Figure~\ref{fig:framework} shows the PR$^{2}$ framework.

\subsection{Problem Formulation}\label{problem}

The training dataset is defined as 
$D = \{(x_i, u_i,  r_i, E_{x_i})\}_{i=1}^{|D|}$, 
where $x_i$ denotes the input question, $u_i = \{d_i\}_{i=1}^{|u_i|}$ represents the associated user profile containing documents about user's previous interactions with the system, $r_i$ is the corresponding user-provided narrative for the question, and $E_{x_i} = \{ e_j \}_{j=1}^{|E_{x_i}|}$ is a set of personalized rubrics that can be used to evaluate generated personal responses for the user. Given $(x_i, u_i)$ to the policy model, the objective is to generate a personalized response $\hat{y}_i$ by leveraging the user-specific information contained in $u_i$. To evaluate $\hat{y}_i$, following~\cite{salemi-zamani-2025-lamp}, we use rubric aspects $E_{x_i}$ and the user narrative $r_i$ that the response to question $x_i$ is expected to address. We then compute a scoring function $\mu(x_i, \hat{y}_i, E_{x_i}, r_i)$ to assess the extent to which $\hat{y}_i$ adequately covers these aspects (the evaluation strategy is explained in Section~\ref{sec:exp-setup}). We model the policy using an LLM $\pi_\theta$ that generates responses conditioned on the question $x_i$ and the user profile $u_i$, which it can access by submitting queries to a retrieval model $\mathcal{R}$. The policy $\pi_\theta$ is initialized from a pretrained LLM checkpoint and optimized under the GRPO \cite{shao2024deepseekmath} reinforcement learning framework. A frozen reference model $\pi_{\text{ref}}$ is used to regularize policy updates.

\subsection{Personalized Rollout Sampling Strategy}\label{sampling}

This section describes the rollout sampling procedure in PR$^{2}$. 
Following Search-R1~\cite{jin2025searchr}, PR$^{2}$ adopts the RAR paradigm, allowing the policy model $\pi_{\theta}$ to interact with an external tool, here a retriever, during rollout generation. For each training instance $(x_i, u_i)$, the current policy samples a group of $G$ personalized trajectories
$\{\hat{y}_i^{(g)}\}_{g=1}^{G} \sim \pi_\theta(\cdot \mid x_i, u_i) \otimes \mathcal{R}$,
where $\otimes$ denotes the interleaving of token generation with retriever invocations (via $R$) to fetch relevant user information from the profile. Thus, for each instance, we obtain a set of personalized sampled trajectories $Y = \{ y_1, \ldots, y_G \}$. The prompt template used for this personalized sampling is: 

\medskip
\begin{adjustwidth}{0.0em}{0.0em} 
\footnotesize
\setstretch{0.} 
\noindent
Your task is to generate a personalized response to the user's question. To do this, you can perform a series of
actions, including thinking in {\color{blue}\texttt{<think>}} and {\color{blue}\texttt{</think>}} tags, searching for information from the user past interactions with the system (i.e., previous asked questions and the detailed information need) by generating a
non-empty search query in {\color{red}\texttt{<search>}} and {\color{red}\texttt{</search>}} tags, and finally providing the answer in {\color{teal}\texttt{<answer>}} and {\color{teal}\texttt{</answer>}} tags. The retrieved information from user history will be provided to you inside {\color{violet}\texttt{<information>}} and {\color{violet}\texttt{</information>}} tags. You need to first think about the question and how to generate a personalized answer for the user. In this thinking process, you should try to understand the user's preferences and needs based on its past interactions with the system. The thinking process should be inside {\color{blue}\texttt{<think>}} and {\color{blue}\texttt{</think>}} tags. If you need to search for information about the
user from its history, you can do this by generating a non-empty search query inside {\color{red}\texttt{<search>}} and {\color{red}\texttt{</search>}} tags. You can use this information in thinking process and answer generation. Nothing should be outside the mentioned tags except the initial question. Now, answer the following question: \texttt{\{question\}}
\end{adjustwidth}

\subsection{Personalized Reward Modeling}\label{Sec:Reward}

The reward function provides the primary supervision signal for policy optimization. In PR$^{2}$, we adopt an outcome-based reward scheme that relies exclusively on the final response rewards to guide the policy model for optimization. Specifically, as described in Section~\S\ref{problem}, for each query–response pair $(x_i, \hat{y}_i)$, an automatic evaluator from \cite{salemi-zamani-2025-lamp} assigns a discrete score based on $\mu(x_i, \hat{y}_i, E_{x_i},r_i)$, measuring the extent to which the response satisfies the user-specific criteria derived from $E_{x_i}$. Formally, the reward is defined as $r(x_i, \hat{y}_i) = \mu(x_i, \hat{y}_i, E_{x_i}, r_i)$. To compute reward, following \cite{salemi-zamani-2025-lamp}, we employ Qwen2.5-32B~\cite{qwen2.5} as the judging model. For each query, the judge assigns a 0–2 score to each personalized aspect based on how well it is addressed. These values are rescaled to the interval $[0,1]$  and averaged across aspects. Additional details can be found in \cite{salemi-zamani-2025-lamp}. These rewards are then used to compute group-relative advantages that drive policy optimization.

\subsection{Advantage Computation}

In GRPO, the advantage is computed as a group-normalized reward for each output~\cite{shao2024deepseekmath}, and policy updates are driven by relative comparisons within a sampled response set. This formulation enables the policy to learn from differences in reward across multiple trajectories generated for the same input. However, prior work has shown that personalization does not consistently yield performance gains, and that non-personalized models can outperform personalized ones in many cases~\cite{10.1145/3626772.3657783}. Consequently, it is desirable to encourage the policy to apply personalization by retrieving from the user profile only when doing so improves performance relative to non-personalized inference. To address this within the GRPO framework, we incorporate a non-personalized baseline when computing the advantage. This enables direct reward-based comparison between retrieval-augmented personalization and non-personal model, allowing the policy to learn when personalization truly improves response quality. 
As a result, the model is encouraged to favor retrieval from profile only when it yields measurable gains over no personalization. A detailed analysis is provided in Section~\S\ref{vanilla}.

To obtain non-personalized responses, we generate a baseline trajectory without retrieval by conditioning only on the input, $\hat{y}_{\text{no-per}} \sim \pi_{\text{ref}}(\cdot \mid x_i)$. Such trajectories involve no retrieval and therefore no personalization. Each non-personalized baseline response $\hat{y}_{\text{no-per}}$ is assigned a reward $r_{\text{no-per}}$ computed as defined in Section~\S\ref{Sec:Reward}. To compute the group-relative advantage under the personalized setting, we first compute group-level statistics over all personalized responses, $\mu = \operatorname{mean}(r_1, \ldots, r_n)$ and $\sigma = \operatorname{std}(r_1, \ldots, r_n)$. For each personalized response $\hat{y}_i \in Y$, the group-relative advantage is defined as $A_i = \frac{r_i - r_{\text{no-per}} - \mu}{\sigma}$. This formulation encourages the model to produce personalized responses when they outperform the non-personalized baseline, and to fall back to non-personalized responses when personalization is detrimental, since the policy can choose to perform no retrieval from the user profile.

\subsection{Summary}
PR$^{2}$ is optimized using GRPO, which performs updates based on relative comparisons among grouped response samples without learning a value function~\cite{shao2024deepseekmath, guo2025deepseek}. 
For each query $x_i$, we sample a group of responses $Y$ as described in Section~\S\ref{sampling}. We then apply the GRPO objective and corresponding policy update to each sampled response $\hat{y}_i \in Y$, following the approach in~\cite{jin2025searchr}. In summary, the contributions of this approach compared to the literature include: (1) developing and evaluating the first reasoning models that retrieve from user profile for personalization, (2) optimizing a policy model for sampling personalization rollouts, (3) applying an aspect-based personalized evaluation to the generated responses as the reward function, and (4) including both personalized and non-personalized trajectories in computing the relative advantage function.



\section{Experiment}
\subsection{Experiment Setup}
\label{sec:exp-setup}

\subsubsection*{\textbf{Datasets \& Evaluation.}}
We conduct experiments on LaMP-QA~\cite{salemi-zamani-2025-lamp}, a benchmark designed for personalized QA. LaMP-QA spans three domains: Art \& Entertainment, Lifestyle \& Personal Development, and Society \& Culture.\footnote{For brevity, in this paper, we refer to \textit{Art \& Entertainment} as ``Art,'' \textit{Lifestyle \& Personal Development} as ``Lifestyle,'' and \textit{Society \& Culture} as ``Society.''} Each instance is accompanied by a user input query and narrative, with the user's historical questions serving as the user profile. For evaluation, following~\cite{salemi-zamani-2025-lamp}, we evaluate responses from a personalized perspective by measuring how well they satisfy user-specific rubrics. We use the instruction-tuned Qwen 2.5-32B~\cite{qwen2.5} with temperature 0.0 as the judge LLM. For each response, the LLM scores each personalized rubric aspect on a scale from 0 to 2. The scores are then normalized to $[0,1]$, and the final score is computed as the average across all aspects. This produces a scalar score measuring alignment with user preferences, enabling fine-grained evaluation of personalized answer quality. For details on the evaluation, we refer the reader to \citet{salemi-zamani-2025-lamp}.

\subsubsection*{\textbf{Baselines.}}
We compare PR$^{2}$ against diverse baselines spanning different modeling perspectives. These include (i) a non-personalized baseline,~\textbf{No Personalization}, following~\cite{salemi-zamani-2025-lamp}, which utilizes CoT to generate responses without incorporating user information, (ii) reasoning-oriented retrieval methods such as \textbf{Search-R1}~\cite{jin2025searchr}, and (iii) personalization-enhanced methods for reasoning and QA, including \textbf{RAG-Personalization} following method in~\cite{salemi-zamani-2025-lamp}, \textbf{PPlug}~\cite{liu-etal-2025-llms}, \textbf{HYDRA}~\cite{zhuang2024hydra}, and \textbf{PrLM}~\cite{10.1145/3746252.3760851}. All baselines share the same retriever, corpus, LLM, temperature (1.0), and token limit (2048).
\begin{table}[t]
\centering
\footnotesize
\caption{Performance on the LAMP-QA test set. * denotes statistically significant improvements compared to all baselines with Bonferroni correction ($p$-value $< 0.01$
).}
\vspace{-1em}
\begin{adjustbox}{width=\columnwidth, center}
\setlength{\tabcolsep}{4pt} 
\renewcommand{\arraystretch}{1.2}
\begin{tabular}{@{}llccc|c@{}} 
\hline
\textbf{Base LLM} & \textbf{Method} & \textbf{Art} & \textbf{Lifestyle} & \textbf{Society} & \textbf{Avg.}\\
\hline
\multirow{7}{*}{Qwen2.5-3B} & No personalization & 0.2503 & 0.3511 & 0.4073 & 0.3362 \\
 & Search R1-GRPO & 0.3411 & 0.4691 & 0.5140 & 0.4414\\
 & RAG-Personalization & 0.2753 & 0.3728 & 0.4282 & 0.3588\\
 & PPlug  & 0.3374 & 0.4522 & 0.5077 & 0.4324\\
 & HYDRA-Adapter & 0.2942 & 0.4504 & 0.5211 & 0.4219\\
 & PrLM & 0.3228 & 0.4762 & 0.5276 & 0.4422\\
\hdashline 
 & \textbf{PR$^{2}$} & \textbf{0.3751*} & \textbf{0.4924*} & \textbf{0.5761*} & \textbf{0.4812*} \\
\hline
\multirow{7}{*}{Qwen2.5-7B} & No personalization & 0.2981 & 0.4014 & 0.4352 & 0.3782\\
 & Search R1-GRPO & 0.3668 & 0.4939 & 0.5416 & 0.4674 \\
 & RAG-Personalization & 0.3086 & 0.4107 & 0.4590 & 0.3928\\
 & PPlug  & 0.3488 & 0.4773 & 0.5301 & 0.4520\\
 & HYDRA-Adapter & 0.3612 & 0.4786 & 0.5319 & 0.4572\\
 & PrLM & 0.3849 & 0.4808 & 0.5422 & 0.4693\\
\hdashline
 & \textbf{PR$^{2}$} & \textbf{0.4265*} & \textbf{0.5519*} & \textbf{0.5983*} & \textbf{0.5256*}\\
\hline
\multirow{7}{*}{Gemma3-4B} & No personalization & 0.2917 & 0.3871 & 0.4462 & 0.3750 \\
 & Search R1-GRPO & 0.3527 & 0.4731 & 0.5266 & 0.4508\\
 & RAG-Personalization & 0.2887 & 0.3965 & 0.4234 & 0.3695 \\
 & PPlug  & 0.3459 & 0.4362 & 0.5141 & 0.4320\\
 & HYDRA-Adapter & 0.3271 & 0.4625 & 0.5273 & 0.4389\\
 & PrLM & 0.3266 & 0.4817 & 0.5301 & 0.4461 \\
\hdashline
 & \textbf{PR$^{2}$} & \textbf{0.3855*} & \textbf{0.5271*} & \textbf{0.5732*} & \textbf{0.4952*}\\
\hline
\end{tabular}
\end{adjustbox}
\label{tab:results}
\vspace{-0.5cm}
\end{table}
\subsubsection*{\textbf{Training Configuration.}}
We conduct experiments using Qwen-2.5-3B, Qwen-2.5-7B, and Gemma3-4B Base models. For retrieval, we employ Contriever~\cite{lei-etal-2023-unsupervised}, a dense retriever fine-tuned on MS MARCO (msmarco-bert-base-dot-v5)~\cite{nguyen2016ms}\footnote{Avilable at:\url{https://hf.co/sentence-transformers/msmarco-bert-base-dot-v5}}, together with FAISS for efficient indexing~\cite{douze2025faiss}. For each query, we retrieve the top $k=3$ from document from user profile; ablations on $k$ are in Section~\S\ref{topk}. Training is performed on the training split of the LaMP-QA benchmark~\cite{salemi-zamani-2025-lamp}. We optimize all models for 200 steps using a fixed rollout budget of 5 per instance, following~\cite{10.1145/3746252.3760851}. We use the vLLM~\footnote{https://github.com/vllm-project/vllm} with temperature 1.0 for efficient response generation and rollout collections. We train with a policy learning rate of $1\times10^{-6}$ using the TRL GRPO implementation\footnote{Available at:~\url{https://hf.co/docs/trl/en/grpo_trainer}}
and a warm-up ratio of 0.285. The KL-divergence regularization coefficient $\beta$ and the clipping parameter $\epsilon$ are fixed to 0.001 and 0.2, respectively, across all experiments. All experiments use NVIDIA A100 GPUs (80GB VRAM, 128GB RAM) with a maximum 2048-token input–output budget.

\section{Results and Discussion}

\subsubsection*{\textbf{Comparison with the Baselines}}
Table~\ref{tab:results} compares PR$^{2}$ with a diverse set of baselines. Across all settings, PR$^{2}$ consistently achieves the highest performance with statistical significance testing with Bonferroni correction~\cite{armstrong2014use}. For Qwen2.5-3B, PR$^{2}$ achieves an average score of 0.4812, outperforming the strongest baseline, PrLM (0.4422), by a 8.8\% relative gain. For Qwen2.5-7B, PR$^{2}$ achieves an average score of 0.5256, outperforming the strongest baseline, PrLM (0.4693), by a 12.0\% relative gain. For Gemma3-4B, PR$^{2}$ reaches an average score of 0.4952, exceeding the strongest baseline, Serach-R1 (0.4508) by a 9.8\% relative gain. These consistent gains demonstrate the effectiveness of PR$^{2}$, which trains the policy to leverage reasoning and retrieval from the user profile to access more useful personal information for each question. A case study example of PR$^{2}$'s output is presented below.

\vspace{1em}
\begin{adjustwidth}{0.2em}{0.2em}
\begin{samepage}
\footnotesize
\setstretch{0.1}
\noindent
\hrule
\vspace{0.4em}
\noindent \textbf{Question}: How many papers will be expected to be published as a postdoc?
\hrule
\vspace{0.4em}

\noindent \textbf{Response:} 
{\color{blue}\texttt{<think>}}
To provide a personalized answer, the specific field of study and institution of the postdoc would be required.
{\color{blue}\texttt{</think>}}

\noindent{\color{red}\texttt{<search>}}
- Field of study, Name of the institution
{\color{red}\texttt{</search>}}

\noindent{\color{violet}\texttt{<information>}}
[{1: 'Incompatibility between job and field of study Finding jobs for Ph.D graduate in
mathematics is the main concern. Available options are namely in general industry and academia. I see, for instance, some Ph.D in math working in Bank sector! This incompatibility between field of study and job might back to the rank of math department. But I want to know about the upper bound of ranks of departments might be considered as top ranks? Additionally, which university ranking is most reliable in industry sector and in academia?} {2: 'What are diplomas, postgraduate courses, specialization, training courses? I am going to apply for some faculty positions.... I have already attended several summer schools, workshops and also took online courses. Can these events be considered as one of the elements of mentioned category?'} {3: 'What is the meaning of pre-doctoral program?... Can someone be a predoctoral in university X and then goes to university to start Ph.D?'}]
{\color{violet}\texttt{</information>}}

\noindent{\color{blue}\texttt{<think>}}
The postdoc’s question about expected publications depends heavily on the field of study and the institution. In this case, the field is mathematics, and there is no information available about the institution.
{\color{blue}\texttt{</think>}}

\noindent{\color{teal}\texttt{<answer>}}
As a postdoc, the number of publications expected can vary significantly depending on both the field and institutional norms. In
mathematics
, a common expectation may be around one to two publications per year, although this can vary widely based on the institution, the complexity of the research, and the pace of progress. It is important to align publication goals with the expectations of the postdoctoral program and the guidance of one’s advisor or mentor to ensure that appropriate benchmarks are being met.
{\color{teal}\texttt{</answer>}}
\hrule
\end{samepage}
\end{adjustwidth}

\begin{table}[t]
\centering
\footnotesize
\caption{Ablation of advantage function in PR$^{2}$.}
\vspace{-0.4cm}
\begin{adjustbox}{width=\columnwidth, center}
\setlength{\tabcolsep}{6pt}
\renewcommand{\arraystretch}{1.2}
\begin{tabular}{@{}llccc|c@{}} 
\hline
\textbf{Model} & \textbf{Method} & \textbf{Art} & \textbf{Lifestyle} & \textbf{Society} & \textbf{Avg.}\\
\hline
\multirow{2}{*}{Qwen2.5-3B} & w. non-personalized rollout  & 0.3751 & 0.4924 & 0.5761 & 0.4812 \\
                            & w.o non-personalized rollout & 0.3345 & 0.4476 & 0.5342 & 0.4387 \\
\hline
\multirow{2}{*}{Qwen2.5-7B} & w. non-personalized rollout  & 0.4265 & 0.5519 & 0.5983 & 0.5256 \\
                            & w.o non-personalized rollout & 0.3706 & 0.4862 & 0.5641 & 0.4736 \\
\hline
\multirow{2}{*}{Gemma3-4B}  & w. non-personalized rollout  & 0.3855 & 0.5271 & 0.5732 & 0.4952 \\
                            & w.o non-personalized rollout & 0.3472 & 0.4793 & 0.5117 & 0.4461 \\
\hline
\end{tabular}
\end{adjustbox}
\label{tab:vanilla_results}
\end{table}

\begin{table}[t]
\centering
\footnotesize
\caption{Retrieved document count on PR$^{2}$ (Qwen2.5-7B).}
\vspace{-0.4cm}
\setlength{\tabcolsep}{7pt}
\renewcommand{\arraystretch}{1.3}
\begin{tabular}{lccc |c}
\hline
\textbf{Method} &
\textbf{Art} &
\textbf{Lifestyle} &
\textbf{Society} &
\textbf{Avg.}\\
\hline
topk=1 & 0.3866 & 0.4468 & 0.5441 & 0.4592 \\
topk=3 & \textbf{0.4265} & \textbf{0.5519} & \textbf{0.5983} & \textbf{0.5256} \\
topk=5 & 0.3918 & 0.4602 & 0.5483 & 0.4668 \\
\hline
\end{tabular}
\label{tab:top}
\vspace{-1.5em}
\end{table}

\subsubsection*{\textbf{Training Dynamics of PR$^{2}$}}

\begin{figure}[t]
    \centering
    \begin{subfigure}[t]{0.15\textwidth}
        \centering
        \includegraphics[width=\linewidth]{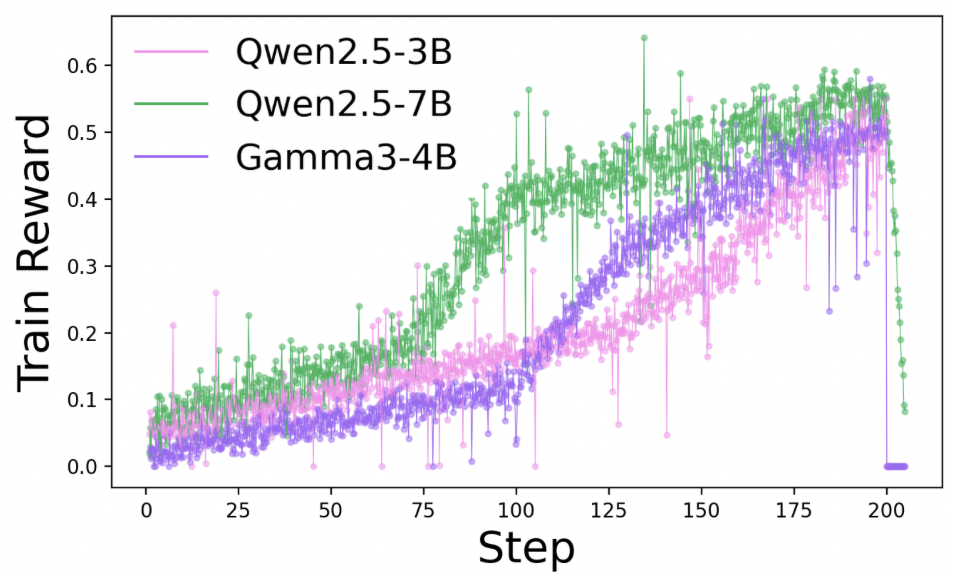}
        \vspace{-2em}
        \caption{Training process}
        \label{Fig:Training process}
    \end{subfigure}
    \hfill
    \begin{subfigure}[t]{0.15\textwidth}
        \centering
        \includegraphics[width=\linewidth]{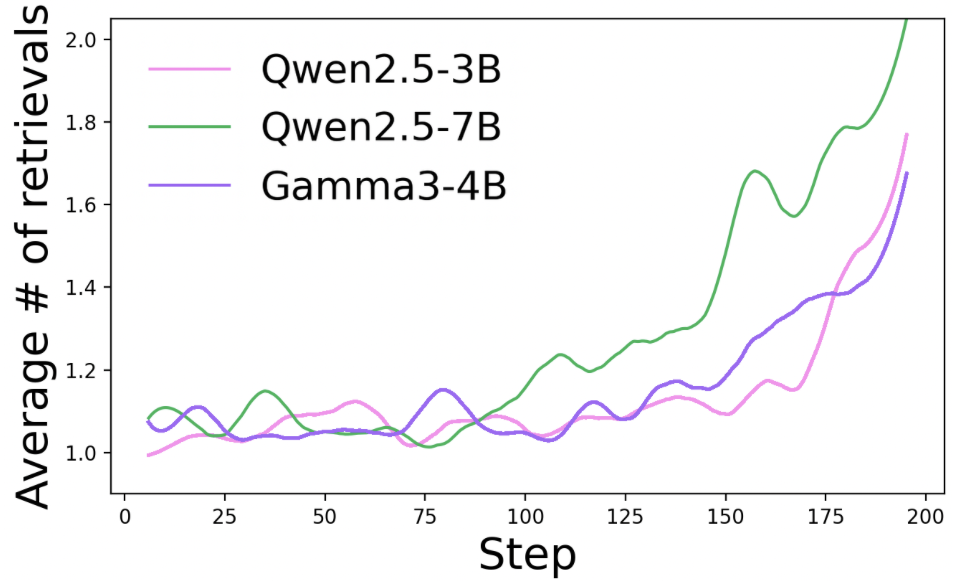}
        \vspace{-2em}
        \caption{Avg \# of retrievals}
        \label{Fig:retrievals}
    \end{subfigure}
    \hfill
    \begin{subfigure}[t]{0.15\textwidth}
        \centering
        \includegraphics[width=\linewidth]{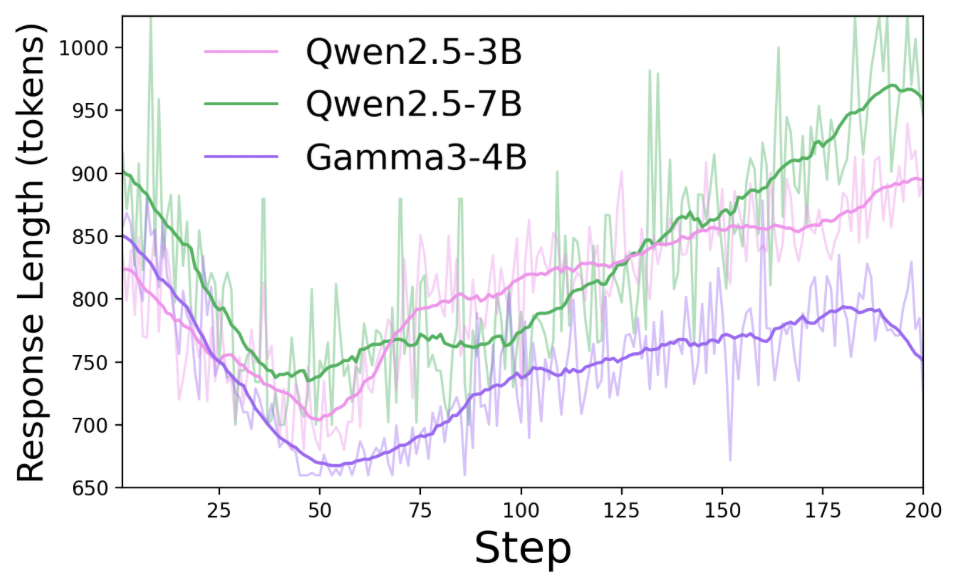}
        \vspace{-2em}
        \caption{Response length}
        \label{Fig:length}
    \end{subfigure}
    \vspace{-1em}
    \caption{The log of reward, response length, and retrieval numbers on the LAMP-QA training.}
    \vspace{-2em}
    \label{fig:three_plots}
\end{figure}

We analyze PR$^{2}$'s GRPO training dynamics with all LLMs. As shown in Figure~\ref{Fig:Training process}, training reward increases steadily for all models, with improvements continuing throughout training and no signs of reward collapse or instability in 200 steps. This trend suggests that GRPO effectively optimizes PR$^{2}$'s policy for personalization-aware retrieval and reasoning, and Qwen2.5-7B consistently reaches the highest reward values, indicating a stronger capacity to leverage retrieved user-specific evidence during reasoning. Figure~\ref{Fig:retrievals} reports the average number of retrieval actions. Early in training, all models perform roughly one retrieval on average, but retrieval frequency increases over time, indicating that the LLM progressively learns to retrieve more frequently as training proceeds. Qwen2.5-7B retrieves the most, whereas Gemma3-4B and Qwen2.5-3B retrieve fewer documents, supporting the interpretation that larger models better exploit additional user-specific retrieval for personalized reasoning. Finally, Figure~\ref{Fig:length} shows a structured progression in response length, exhibiting a decrease–increase–stabilize pattern. During the first 100 steps, response length decreases while reward improves. After the first 100 steps, response length increases with reward gains, indicating the LLM more frequently invokes the retriever and incorporates retrieved content. The rising reward suggests that the model becomes more effective at leveraging retrieved content to personalize answers. Across models, Qwen2.5-7B produces the longest responses, 3B the shortest, with Gemma3-4B in between.

\subsubsection*{\textbf{Effect of Non-personalized Baseline in Advantage Computation} \label{vanilla}}

Table~\ref{tab:vanilla_results} examines the role of including non-personalized baseline in training. Removing the baseline reduces performance: Qwen2.5-3B drops from 0.4812 to 0.4387, Qwen2.5-7B from 0.5256 to 0.4736, and Gemma3-4B from 0.4952 to 0.4461. These results show that incorporating non-personalized baseline alongside personalized sampling consistently improves performance, showing that they are a critical component of PR$^{2}$'s learning process. By contrasting personalized and non-personalized trajectories under the same reward objective, the model learns when retrieval meaningfully enhances personalization rather than overusing retrieval or relying on noisy personalized signals. This contrast allows PR$^{2}$ to better attribute improvements to the integration of retrieved evidence rather than to generic generation gains. Without it, the policy is less capable of distinguishing when retrieval from profile improves generation, resulting in weaker learned behavior.

\subsubsection*{\textbf{Effect of the Number of Retrieved User-Profile Documents} \label{topk}}

We analyze how the number of retrieved user-profile documents affects PR$^{2}$. Following~\cite{10.1145/3746252.3760851}, our main setting uses top-k = 3; we also evaluate top-k = 1 and 5. Results on Qwen2.5-7B show that top-k = 3 achieves the best overall performance. The findings reveal a trade-off between retrieval sufficiency and noise: retrieving few documents likely limits recall of relevant user-profile information, constraining the LLM's ability to fully capture personalized nuances in profile, while retrieving too many may introduce irrelevant documents that reduce precision and interfere with reasoning. 

\section{Conclusion}

This paper introduces PR$^{2}$, an RL framework that integrates retrieval directly into the reasoning process for personalized QA. Unlike prior studies that focus primarily on retrieval without explicitly modeling the reasoning process, PR$^{2}$ jointly optimizes retrieval and reasoning decisions under personalized reward signals. By training with GRPO and contrasting personalized and non-personalized rollouts in advantage calculation, the framework learns when retrieval is necessary and how retrieved user-specific evidence should shape reasoning steps. Experiments on LaMP-QA show consistent gains across multiple LLMs, demonstrating that aligning retrieval and reasoning with user-specific objectives improves personalized QA.
\newline
\newline
\noindent
{\bf Acknowledgment}
\newline
\noindent
This work was supported in part by the Center for Intelligent Information Retrieval, in part by NSF grant \#2402873, in part by the Office of Naval Research contract \#N000142412612, and with support from by Google.org. Any opinions, findings and conclusions or recommendations expressed in this material are those of the authors and do not necessarily reflect those of the sponsor.

\bibliographystyle{ACM-Reference-Format}
\balance
\bibliography{ref}


\end{document}